\documentclass{article}

\usepackage[preprint]{neurips_2025}

\usepackage[utf8]{inputenc} 
\usepackage[T1]{fontenc}    
\usepackage{hyperref}       
\usepackage{url}            
\usepackage{booktabs}       
\usepackage{amsfonts}       
\usepackage{nicefrac}       
\usepackage{microtype}      
\usepackage{xcolor}         
\usepackage{subcaption}   
\usepackage{graphicx}
\usepackage{tcolorbox}     
\usepackage{algorithm,algpseudocode}
\usepackage{amsmath} 

\title{CodeMixBench: Evaluating Large Language Models on Code Generation with Code-Mixed Prompts}

\author{%
  Manik Sheokand \\
  Department of Computer Science \\
  Chandigarh University \\
  Punjab, India \\
  \texttt{21bcs11856@cuchd.in}
  \And
  Parth Sawant \\
  Department of Technology Management and Innovation \\
  New York University \\
  New York, USA \\
  \texttt{ps5624@nyu.edu}
}

\begin{document}

\maketitle


\begin{abstract}
Large Language Models (LLMs) have achieved remarkable success in code generation tasks, powering various applications like code completion, debugging, and programming assistance. However, existing benchmarks such as HumanEval, MBPP, and BigCodeBench primarily evaluate LLMs on English-only prompts, overlooking the real-world scenario where multilingual developers often use code-mixed language while interacting with LLMs. To address this gap, we introduce \textit{CodeMixBench}, a novel benchmark designed to evaluate the robustness of LLMs on code generation from code-mixed prompts. Built upon BigCodeBench, CodeMixBench introduces controlled code-mixing (CMD) into the natural language parts of prompts across three language pairs: Hinglish (Hindi-English), Spanish-English, and Chinese Pinyin-English. We comprehensively evaluate a diverse set of open-source code generation models ranging from 1.5B to 15B parameters. Our results show that code-mixed prompts consistently degrade Pass@1 performance compared to their English-only counterparts, with performance drops increasing under higher CMD levels for smaller models. CodeMixBench provides a realistic evaluation framework for studying multilingual code generation and highlights new challenges and directions for building robust code generation models that generalize well across diverse linguistic settings.
\end{abstract}

\section{Introduction}

Large Language Models (LLMs) have revolutionized code generation~\cite{bielik2016phog, yin2017syntactic}, enabling tasks such as automated synthesis, completion, bug fixing, and summarization. Models like Codex~\cite{chen2021evaluating}, AlphaCode~\cite{li2022alphacode}, StarCoder~\cite{li2023starcoder}, and DeepSeek-Coder~\cite{deepseek2024coder} have achieved strong results by training on large-scale code datasets. However, these advancements typically assume monolingual (English-only) interactions between developers and LLMs—an assumption that does not reflect the reality of global software development.

In practice, developers frequently blend their native language with English when writing comments, docstrings, or giving natural language instructions—a phenomenon known as code-mixing or code-switching~\cite{myers1993social, solorio2008learning, bali2014borrowing}. This behavior is common in multilingual communities, where using native expressions alongside English code constructs is both natural and intuitive. Despite the increasing adoption of LLMs, their ability to handle such code-mixed prompts remains underexplored.

Existing code generation benchmarks such as HumanEval~\cite{chen2021evaluating}, MBPP~\cite{austin2021program}, APPS~\cite{hendrycks2021apps}, and BigCodeBench~\cite{bigcode2024bench} evaluate models using only English prompts, failing to account for the linguistic diversity found in real-world programming.

To address this gap, we introduce \textit{CodeMixBench}, a benchmark designed to evaluate LLMs on code generation tasks where prompts are written in a code-mixed format combining English with another language. Built on top of BigCodeBench, CodeMixBench utilizes the concept of controllable code-mixing degree (CMD) introduced by~\cite{gupta2024controlled} spanning English-Hindi (Hinglish), Spanish-English, and Chinese Pinyin-English. We employ an LLM-driven augmentation pipeline using Gemini-2.0-Flash-Lite for translation and Controlled Generation (CG)~\cite{gupta2024controlled} to modulate code-mixing while preserving task semantics.

We evaluate a diverse set of open-source code generation models (1.5B–15B parameters) across CMD levels and languages. Our results show that code-mixing consistently degrades performance, especially at high CMD values; larger and instruction-tuned models like OpenCoder-8B-Instruct~\cite{opencoder2024} and DeepSeek-Coder~\cite{deepseek2024coder} are more robust, while smaller models suffer significantly. CodeMixBench thus provides a new lens into the multilingual generalization capabilities of code LLMs and highlights the importance of evaluating model robustness beyond monolingual assumptions.

\section{Related Work}
\label{related_work}

\subsection{Code Generation Benchmarks}

Several benchmarks have been proposed to evaluate the code generation capabilities of LLMs, primarily using English-only prompts. HumanEval~\cite{chen2021evaluating} introduced a set of hand-crafted Python tasks focused on functional correctness, while MBPP~\cite{austin2021program} extended this with 974 entry-level tasks accompanied by unit tests. APPS~\cite{hendrycks2021apps} contributed a large-scale benchmark with 10,000 problems from competitive programming platforms, and xCodeEval~\cite{xu2022xcodeeval} emphasized multilingual executable code across 11 programming languages.

BigCodeBench~\cite{bigcode2024bench} builds on these efforts with 1,140 realistic tasks requiring function calls from 139 Python libraries across diverse domains. It emphasizes tool use and compositional reasoning, requiring multi-step synthesis and validation via high-coverage unit tests. However, like its predecessors, BigCodeBench assumes monolingual (English-only) instructions and does not address LLM robustness to multilingual or code-mixed prompts.

\subsection{Multilingual Code Models and Code-Mixing}

Recent LLMs such as CodeLLaMA~\cite{roziere2023codellama}, StarCoder2~\cite{li2023starcoder}, and DeepSeek-Coder~\cite{deepseek2024coder} have introduced multilingual support for code understanding and generation. These models are typically trained on datasets containing multiple programming languages and are evaluated on tasks such as completion and translation. However, their focus remains on code syntax diversity rather than handling code-mixed natural language instructions embedded within prompts.

In parallel, code-mixing has been extensively studied in NLP for tasks such as text classification~\cite{chaturanga2021classification}, translation~\cite{vavre2022translation}, and sentiment analysis~\cite{perera2024sentiment}. Data augmentation techniques such as controlled generation~\cite{gupta2024controlled}, and semantic evaluation methods like Gold-standard Agnostic Measure for Evaluation (GAME)~\cite{gupta2024controlled} and Metric-Independent Pipeline for Evaluation (MIPE)~\cite{garg2021mipe}, have advanced the quality of code-mixed datasets and their evaluation. However, code-mixing in the context of natural language-to-code generation remains underexplored, particularly in relation to preserving task semantics and ensuring executable correctness.

\subsection{Evaluation Metrics for Code Generation}

Pass@k~\cite{chen2021evaluating} has become the standard metric for evaluating the functional correctness of generated code, measuring the likelihood that at least one of the top-$k$ outputs passes all associated test cases. While suitable for execution-based validation, it does not capture whether code-mixed prompts retain their intended meaning. Metrics like BLEU~\cite{papineni2002bleu} are poorly suited for this purpose due to lexical variability and the absence of gold-standard references in code-mixed text. To address this, we adopt the GAME score, an embedding-based metric designed for code-mixed evaluation. GAME compares sentence embeddings of back-translated prompts to the original using cosine similarity, providing a semantic fidelity score on a 0–100 scale.

Our work is the first to integrate these threads — multilingual NLP, code generation benchmarks, and semantic evaluation — into a unified benchmark that systematically evaluates code generation LLMs under code-mixed prompt conditions.

\section{Dataset: CodeMixBench}
\label{dataset}

CodeMixBench is constructed as an augmentation of BigCodeBench~\cite{bigcode2024bench}, a benchmark for complex function-level code generation. BigCodeBench comprises two splits: \textit{complete split}, containing full docstrings for function synthesis, and \textit{instruct split}, containing concise task instructions for instruction-tuned models. We retain both splits and introduce multilingual variations by modifying only the natural language components—prompt instructions and docstrings—while preserving the executable code structure.
Table~\ref{tab:prompt_splits} (see Appendix ~\ref{appendix:prompt_formats_dataset}) shows examples of each.

\subsection{Target Languages and Code-Mixing Scope}

We support three realistic code-mixing configurations based on prevalence in multilingual developer communities: Hinglish (Hindi-English), Spanish-English, and Chinese Pinyin-English. Table~\ref{tab:lang_scope} summarizes the rationale for their inclusion. Code-mixing is applied only to the prompt instructions and docstrings, leaving code unchanged.

\begin{table}[htbp]
\caption{Target language mixes and motivation for inclusion in CodeMixBench.}
\centering
\begin{tabular}{p{0.28\linewidth} p{0.65\linewidth}}
\toprule
\textbf{Language mix} & \textbf{Rationale} \\
\midrule
Hinglish (Hindi-English) & Common in South Asia, especially India; widely observed in informal and technical communication. \\
Spanish-English & Prevalent in Latin America and the US; large bilingual developer base. \\
Chinese Pinyin-English & Reflects common use in East Asia and among Chinese developers using romanized Chinese in prompts. \\
\bottomrule
\end{tabular}
\label{tab:lang_scope}
\end{table}

\subsection{Code-Mixing Methodology}

We adopt a two-stage augmentation strategy: (1) translation to a matrix language (e.g., Hindi) using Gemini-2.0-Flash-Lite while preserving programming tokens, and (2) controlled code-mixing using a CMD $\in [0, 1]$ to determine the proportion of embedded English tokens retained. For example, CMD = 0.6 retains ~60\% of switchable words in English, while CMD = 0.9 retains more. Details of this procedure are shown in the example prompt in Table~\ref{tab:hinglish_example} in Appendix~\ref{appendix:code_mix_example}.

\subsubsection{Implementation Pipeline for CodeMixBench}

The complete pipeline used to construct CodeMixBench is composed of the following four components:

\paragraph{1. Base Translation and Word Dictionary Construction}

We begin by prompting \textit{Gemini-2.0-Flash-Lite} to translate both the \texttt{instruct\_prompt} and the doc string(\texttt{doc\_struct}) into the target matrix language (e.g., Hindi), while preserving programming-specific tokens such as \texttt{args} and \texttt{list}. To identify potential switch points for controlled code-mixing, we apply Part-of-Speech (PoS) tagging~\cite{toutanova2003feature} to the original English sentence using spaCy. This enables us to extract a filtered list of content words—typically nouns, adjectives, and verbs—that are syntactically appropriate for substitution. Using the same LLM prompt, Gemini is then asked to identify the corresponding words in the translated matrix-language sentence for each selected English token. The model returns a bilingual dictionary of aligned word pairs:
\[
W = \{(w_i^{\text{eng}}, w_i^{\text{mtx}})\}_{i=1}^n
\]
where $w_i^{\text{eng}}$ is an English word and $w_i^{\text{mtx}}$ is its matrix-language equivalent. Additionally, for each $w_i^{\text{mtx}}$, Gemini provides three romanized transliterations to capture spelling variability found in real-world code-mixed usage:
\[
R_i = \{\text{roman}_{i}^{(1)}, \text{roman}_{i}^{(2)}, \text{roman}_{i}^{(3)}\}.
\]
This romanization is necessary to simulate informal orthographic patterns prevalent in platforms like Twitter, where Hinglish is often written with inconsistent spellings. As a result, this translation step produces two key artifacts: (1) a fully translated matrix-language version of the prompt with code-related tokens intact, and (2) a structured dictionary mapping
\[
\text{English} \rightarrow \text{Matrix Language} \rightarrow \text{Romanized Forms},
\]

\paragraph{2. Controlled Code-Mix Injection}

To simulate realistic patterns of code-mixing, we apply a replacement strategy based on lexical trends observed in real-world code-mixed text. Specifically, we use a code-mixed Twitter corpus~\cite{nayak-joshi-2022-l3cube} to estimate the frequency of both English words and their Romanized Hindi variants.

For each replaceable word $w_i$, we calculate a replacement score $s_i$ that captures how likely it is for the Hindi variant to be substituted back into English. This score is defined as:
\[
s_i = \frac{f(\text{eng}_i)}{f(\text{hi}_i)}, \quad f(\text{hi}_i) = \sum_{j=1}^{3} f(\text{roman}_{i}^{(j)})
\]
where $f(\cdot)$ denotes frequency in the corpus, and the denominator aggregates the frequencies of up to three Romanized variations for the Hindi translation of $w_i$. A higher score implies that the English form is more dominant or natural in real-world code-mixed usage. If $f(\text{hi}_i) = 0$, we assign $s_i = \infty$, indicating a preference to retain the English word due to the rarity or unnaturalness of its Hindi form.

Given a specified CMD $\in [0, 1]$, we calculate the number of words to be replaced using $\text{floor}(N \cdot \text{CMD})$, where $N$ is the number of eligible switch points—i.e., content words (typically nouns, verbs, and adjectives) identified via PoS tagging in the original English sentence. We then sort the candidate words by their score $s_i$ in descending order and select the top words for replacement. Words with infinite scores are always prioritized and replaced first, even if they exceed the quota implied by the CMD value.

For example, with CMD = 0.7 and three replaceable words, we replace only two of them.\[3 \times 0.7 \approx 2\]

\paragraph{3. Romanization}

After injecting the desired level of code-mixing, we apply a Romanization step to convert any remaining Hindi tokens into the Roman script. This is accomplished via a follow-up prompt to the same LLM, which is instructed to romanize only the Hindi words while leaving programming-specific tokens, punctuation, and English subwords untouched. This selective Romanization ensures that the resulting code-mixed prompts remain structurally intact and syntactically parsable, while being fully compatible with LLMs trained predominantly on Roman script input.

\paragraph{4. Docstring Replacement for Complete Prompts}

For prompts in the \texttt{complete} split, we apply the same translation and controlled code-mixing to the function-level docstring. The original English docstring is first translated into the matrix language and then processed using PoS tagging, frequency-based scoring, CMD-controlled replacement, and Romanization — exactly as described in earlier steps. The final code-mixed and romanized version is returned as a structured dictionary and injected back into the source code, replacing the default English docstring.

\subsection{Semantic Quality Verification: GAME Score}

To evaluate whether code-mixed prompts preserve semantic intent, we use the \textbf{GAME} score. Each prompt is back-translated to English using Gemini-2.0-Flash-Lite, and compared to the original using cosine similarity of sentence embeddings (via all-MiniLM-L6-v2). Cosine similarity is clipped to $[0, 1]$ and scaled to a final GAME score in $[0, 100]$. Across CMD levels 0.6 and 0.9, we observe a mean GAME score of \textbf{90\%}, confirming high semantic fidelity in the generated prompts.

\subsection{Dataset Statistics}

CodeMixBench includes 1,140 unique tasks, each augmented across three language mixes and two CMD levels (0.6, 0.9), using both \texttt{instruct} and \texttt{complete} variants. This results in 6,840 total prompts (3 languages × 2 CMDs × 1140). Table~\ref{tab:dataset_stats} summarizes the distribution.

\begin{table}[htbp]
\caption{Codemixbench example counts across languages and CMD levels.}
\label{tab:dataset_stats}
\centering
\begin{tabular}{lcc}
\toprule
\textbf{Language mix} & \textbf{CMD = 0.6} & \textbf{CMD = 0.9} \\
\midrule
Hinglish & 1140 & 1140 \\
Spanish-English & 1140 & 1140 \\
Pinyin-English & 1140 & 1140 \\
\midrule
\textbf{Total} & 3420 & 3420 \\
\bottomrule
\end{tabular}
\end{table}

Detailed base translation, romanization and all are giving under Appendix~\ref{appendix:prompt-templates} and ~\ref{alg:game-validation}. Code-mix algorithm is given under Appendix~\ref{appendix:cmd-algorithm}. All data, preprocessing scripts, and evaluation tools will be released on Hugging Face and GitHub under a CC-BY 4.0 license upon acceptance.

\section{Experimental Setup}
\label{experiments}

All experiments were conducted using a modified version of the official BigCodeBench harness. Code was executed inside isolated E2B sandbox containers (Python 3.10, preinstalled libraries). Inference ran on a cloud instance hosted by Lightning AI with 16 vCPUs, 128 GB RAM, and a single NVIDIA L40s GPU (48 GB VRAM). Code execution was CPU-based. Full evaluation per model on our dataset completed in under one hour.

\subsection{Models Evaluated}

We evaluate 17 open-source code generation models ranging from 1B to 15B parameters, including instruction-tuned, distilled, and multilingual variants. These span multiple architectures and training objectives (e.g., Qwen2.5, DeepSeek, StarCoder2, OpenCoder, Phi-4). See Appendix~\ref{appendix:model_list} (Table~\ref{tab:model_list}) for the full list of models and parameter sizes.

All models were evaluated in a zero-shot setting using standardized prompts. For \texttt{complete} tasks, we supplied full docstrings; for \texttt{instruct} tasks, concise natural instructions were used. We adopted the default prompting technique provided by the BigCodeBench library to ensure consistency with prior evaluation protocols. Prompt examples across CMD values are provided in Appendix~\ref{appendix:prompt_formats_cmd} (Table~\ref{tab:cmd_prompt_examples}).

\subsection{Evaluation Metric: Pass@1}

We adopt the standard \textbf{Pass@1} metric~\cite{chen2021evaluating}, which checks whether the top-1 generated solution passes all associated test cases. Each model-task pair uses greedy decoding to produce a single code snippet, which is then executed securely in the E2B sandbox and evaluated against BigCodeBench's test suite.

\section{Results and Analysis}
\label{results}

We report Pass@1 performance across English-only prompts (baseline), and code-mixed prompts at CMD = 0.6 (light mixing) and CMD = 0.9 (heavy mixing). Models are evaluated on the \texttt{complete} split of CodeMixBench, with code-mixed prompts evaluated exclusively in the Hindi-English code-mix subset.

\subsection{Overall Pass@1 Performance}

\begin{table}[htbp]
\caption{Pass@1 performance on the \texttt{complete} split across English and code-mixed prompts.}
\centering
\begin{tabular}{lccc}
\toprule
\textbf{Model} & \textbf{English (\%)} & \textbf{CMD = 0.6 (\%)} & \textbf{CMD = 0.9 (\%)} \\
\midrule
DeepSeek-R1-Distill-Qwen-1.5B & 7.9 & 4.1 & 4.8 \\
StarCoder2-3b & 21.4 & 9.3 & 9.1 \\
Llama-3.2-1B & 11.3 & 5 & 4.5 \\
Qwen2.5-Coder-1.5B-Instruct & 32.7 & 31.8 & 22.2 \\
Gemma-3-4b-it & 37.8 & 28.2 & 28.4 \\
OpenCoder-8B-Instruct & 50.9 & 51.3 & 39.0 \\
Phi-4-multimodal-instruct & 46.5 & 46.4 & 33.1 \\
DeepSeek-R1-Distill-Llama-8B & 15.3 & 15.9 & 11.1 \\
Hermes-2-Theta-Llama-3-8B & 36.4 & 36.4 & 29.1 \\
CodeLlama-7b-Instruct-hf & 25.0 & 17.7 & 17.7 \\
Qwen2.5-Coder-7B-Instruct & 48.8 & 41.2 & 41.3 \\
Llama-3.1-8B-Instruct & 40.5 & 38.8 & 31.4 \\
StarCoder2-7b & 27.7 & 6.8 & 8.9 \\
Phi-4 & 55.4 & 46.7 & 47.1 \\
DeepSeek-R1-Distill-Qwen-14B & 48.4 & 38.8 & 40.3 \\
DeepSeek-Coder-V2-Lite-Instruct & 47.6 & 37.7 & 38.1 \\
StarCoder2-15b-instruct-v0.1 & 45.1 & 33.1 & 32.7 \\
\bottomrule
\end{tabular}
\label{tab:overall_results}
\end{table}

Code-mixed prompts consistently reduce Pass@1 performance, with more severe degradation observed at CMD = 0.9. Larger, instruction-tuned models like OpenCoder-8B-Instruct and Phi-4 maintain strong performance across settings, while smaller models (e.g., LLaMA-3.2-1B, StarCoder2-3B) degrade sharply under heavy mixing.

\subsection{Analysis of CMD Level, Language, and Model Size}

\begin{figure}[htbp]
  \centering
  \includegraphics[width=\textwidth]{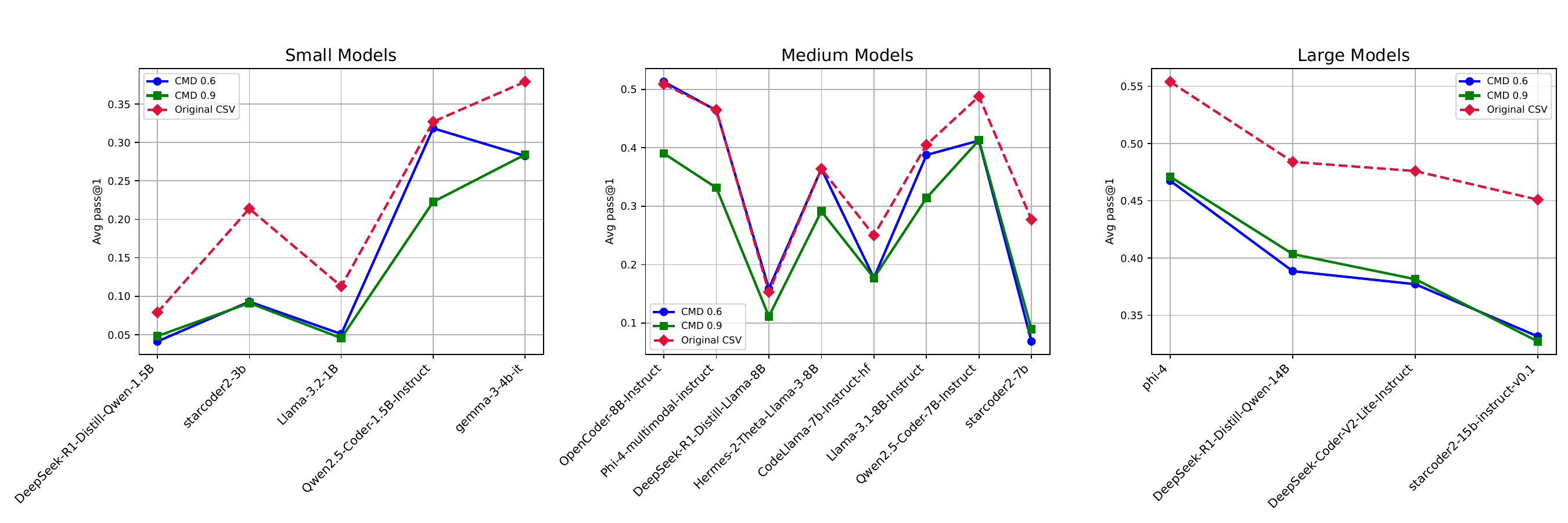}
  \caption{Comparison of CMD 0.6, 0.9 and original English prompt values for pass@1}
  \label{fig:cmd_grid}
\end{figure}

\begin{figure}[htbp]
  \centering
  \includegraphics[width=\textwidth]{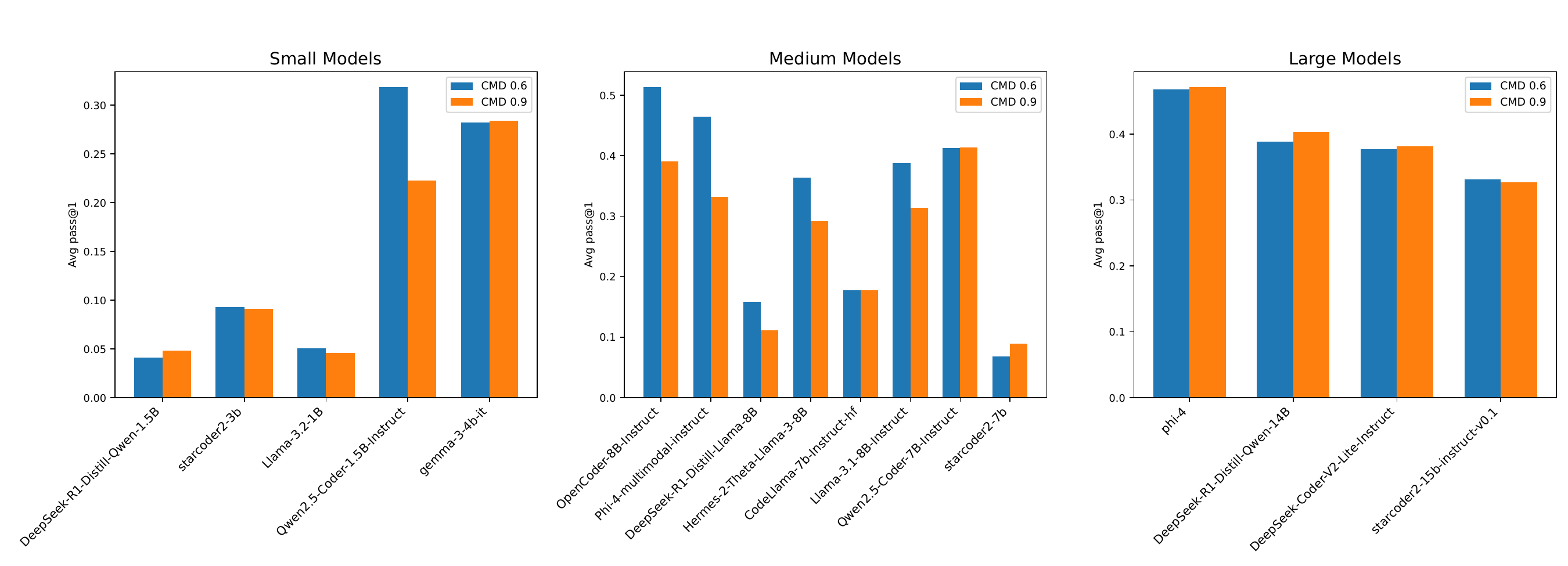}
  \caption{Comparison of CMD 0.6, 0.9 and model size for pass@1}
  \label{fig:cmd_combined}
\end{figure}

Despite the general downward trend, a few models stand out for their robustness to increasing CMD. In particular, \texttt{OpenCoder-8B-Instruct}, \texttt{DeepSeek-R1-Distill-Llama-8B}, and \texttt{Qwen2.5-Coder-1.5B-Instruct} retain near-baseline performance even under heavy code-mixing. \texttt{OpenCoder-8B}~\cite{opencoder2024} exhibits virtually no loss at CMD~0.6 (51.3\% Pass@1 vs. 50.9\% on English). This model’s resilience can be attributed to its training on a massive and diverse dataset (2.5 trillion tokens, 90\% code and 10\% code-related web text) across languages, combined with strong instruction-tuning. Indeed, the \textit{CodeMixBench} findings note that \texttt{OpenCoder-8B}’s ``nearly identical performance on English and code-mixed prompts'' likely due to its multilingual, noisy pretraining and robust fine-tuning regimen.

Similarly, \texttt{DeepSeek-R1-Distill-Llama-8B}~\cite{deepseekai2025deepseekr1incentivizingreasoningcapability} shows only minor degradation as CMD increases. This 8B model maintains high Pass@1 across CMD levels in the medium sized model bracket. Its robustness is shown in the \textit{DeepSeek-Coder} training strategy and distillation technique. The DeepSeek models was trained from scratch on 2 trillion tokens of data comprising 87\% code and 13\% code-mixed natural language. The \texttt{R1-Distill Llama-8B} model in particular was obtained by knowledge distillation from a larger DeepSeek model onto a Llama-based 8B architecture.

\texttt{Qwen2.5-Coder-1.5B-Instruct} is a particularly noteworthy case of a smaller model remaining robust to code-mixing. Despite having only 1.5B parameters, \texttt{Qwen2.5-Coder} shows only a modest drop in Pass@1 when moving to CMD~0.9, especially compared to other models in the 1--3B range (which often collapse under heavy mixing). We attribute \texttt{Qwen2.5}’s strong performance to its highly effective training. It is built on Alibaba’s \textit{Qwen2.5}~\cite{hui2024qwen25codertechnicalreport} architecture and was continued-pretrained on an enormous 5.5 trillion token code corpus. This scale of training far exceeds that of most models in its size class.

\section{Discussion}
\label{discussion}

\subsection{Key Findings and Takeaways}

Our study presents the first systematic benchmark evaluating the robustness of large language models (LLMs) to code-mixed prompts in code generation. Through comprehensive experimentation across multiple models, languages, and levels of code-mixing, we identify several key takeaways. First, code-mixed prompts consistently degrade Pass@1 performance, with higher CMD levels (e.g., CMD = 0.9) producing the most severe drops. Model size and instruction tuning significantly influence robustness, with larger models (7B+) demonstrating greater resilience than smaller models (1B–3B), and instruction-tuned variants consistently outperforming their base counterparts. Notably, OpenCoder-8B-Instruct emerged as exceptionally robust, with minimal performance degradation across all code-mixed settings—highlighting the impact of multilingual and noisy pretraining. Furthermore, our results indicate that models trained on large-scale, diverse corpora—particularly those containing noisy and multilingual natural language—consistently perform better under code-mixed prompts. For instance, Qwen2.5-Coder-1.5B and DeepSeek-R1-Distill-Llama-8B both exhibit strong Pass@1 accuracy despite their smaller or distilled sizes, owing to the breadth and diversity of their training datasets.

\subsection{Why Does Performance Drop on Code-Mixing?}

We hypothesize that several factors contribute to performance degradation under code-mixed conditions. A primary reason is training data bias: most LLMs are trained on English-dominant code datasets, with minimal exposure to multilingual or code-mixed content. Tokenizer limitations also play a role—non-English tokens, especially Romanized text, inflate sequence lengths and lead to fragmented tokenization. Additionally, semantic ambiguity in code-mixed prompts can make it harder for models to ground instructions, and frequent mixing breaks the language modeling priors on which these models rely.

\subsection{Practical Implications}

These findings have practical implications for LLM deployment in multilingual developer environments. Without explicit multilingual or code-mixed pretraining, LLM-based assistants may underperform for non-English or mixed-language users. Instruction tuning alone appears insufficient; instead, dataset diversification with noisy or code-mixed content is necessary to build more inclusive and robust systems. Developers and LLM providers alike should consider refining tokenizer strategies and incorporating multilingual data to better support global usage patterns.

\subsection{Limitations of Our Study}

Despite its contributions, CodeMixBench has several limitations. It currently focuses on three language pairs—Hinglish, Spanish-English, and Chinese-English—leaving many others unexplored. Our code-mixing is applied only to natural language instructions and docstrings, whereas real-world mixing often occurs in comments, variable names, and inline annotations. Furthermore, we focus solely on functional correctness (Pass@1), without evaluating code quality, readability, or adherence to stylistic conventions. Additionally, due to computational and resource constraints, we were limited in the number and variety of models evaluated across multiple code-mixing degrees and language sets. As a result, several models and code-mixing scenarios remain unexplored in this study.

\subsection{Future Work Directions}

Future research can extend CodeMixBench in several directions. Expanding to additional language pairs such as Tamil-English, Bengali-English, and Korean-English would offer broader multilingual coverage. Enhancing realism by introducing code-mixing in comments, variable names, and inline explanations would further reflect developer behavior. Constructing human-authored code-mixed datasets would help eliminate translation artifacts and improve benchmark authenticity. Future work can also explore multilingual-aware tokenizers optimized for Romanized non-English words, and assess the impact of fine-tuning or instruction-tuning on code-mixed generation. Finally, incorporating new metrics—such as code quality, runtime efficiency, and maintainability—would provide a more comprehensive evaluation of model performance in multilingual coding environments.

\section{Conclusion}
\label{conclusion}

In this work, we introduced \textbf{CodeMixBench}, a novel benchmark for evaluating code generation capabilities of large language models (LLMs) under realistic code-mixed prompting scenarios. Built as a controlled augmentation of BigCodeBench, CodeMixBench incorporates graded code-mixing (CMD) across three high-impact multilingual pairs—Hinglish (Hindi-English), Spanish-English, and Chinese Pinyin-English—while ensuring the preservation of functional correctness and task semantics.

Our evaluation of 17 open-source models, spanning parameter sizes from 1.5B to 15B, reveals a consistent degradation in Pass@1 performance with increasing code-mixing, particularly at higher CMD levels (e.g., CMD = 0.9). We find that smaller and base models are disproportionately affected, often failing to generalize in heavily code-mixed settings. In contrast, instruction-tuned and larger models demonstrate greater robustness, though not without limitations. Crucially, we observe that model resilience is not solely a function of scale—models such as Qwen2.5-Coder-1.5B and DeepSeek-R1-Distill-Llama-8B outperform larger counterparts due to their training on diverse, multilingual, and noisy corpora.

CodeMixBench serves as a critical step toward closing the gap between monolingual code benchmarks and the real-world multilingual interactions of software developers. As global software engineering increasingly incorporates code-mixed communication, our benchmark provides the first rigorous framework to assess and improve LLMs under these settings.

All datasets, augmentation scripts, and evaluation pipelines will be released publicly to ensure transparency, reproducibility, and continued progress in this emerging area of multilingual code intelligence.


\section{Appendix}

\subsection{Full list of evaluated models}
\label{appendix:model_list}

\begin{table}[h]
\caption{Full list of evaluated models and their parameter sizes.}
\label{tab:model_list}
\centering
\renewcommand{\arraystretch}{1.1}
\begin{tabular}{p{0.6\linewidth} p{0.25\linewidth}}
\toprule
\textbf{Model name} & \textbf{Size (B)} \\
\midrule
StarCoder2 (base) & 3B, 7B, 15B \\
Qwen2.5-Coder (instruct) & 1.5B, 7B \\
DeepSeek-Coder-V2 (lite-instruct) & 14B \\
DeepSeek-R1-Distill (Qwen, LLaMA variants) & 1.5B, 8B, 14B \\
CodeLLaMA (instruct) & 7B \\
Phi-4 / Phi-4-multimodal & 7B, 14B \\
LLaMA-3.1 / LLaMA-3.2 (base) & 1B, 8B \\
OpenCoder-8B-Instruct & 8B \\
Gemma-3 & 4B \\
Hermes-2-Theta (LLaMA3) & 8B \\
\bottomrule
\end{tabular}
\end{table}

\subsection{CMD 0.9 Code-Mixed prompt example (Hinglish)}
\label{appendix:code_mix_example}

\begin{table}[h]
\caption{Comparison of original English and code-mixed Hinglish prompt at CMD = 0.9.}
\label{tab:hinglish_example}
\centering
\renewcommand{\arraystretch}{1.1}
\begin{tabular}{p{0.47\linewidth} p{0.47\linewidth}}
\toprule
\textbf{Original English prompt} & \textbf{Hinglish prompt (CMD = 0.9)} \\
\midrule
\begin{minipage}[t]{\linewidth}
\footnotesize\ttfamily\raggedright
""" \par
Given a list of numbers, compute the average sum of absolute differences across all permutations. \par
Args: \par
\hspace{1em}numbers (list): List of integers. \par
Returns: \par
\hspace{1em}float: The average sum of differences. \par
"""
\end{minipage}
&
\begin{minipage}[t]{\linewidth}
\footnotesize\ttfamily\raggedright
""" \par
Diye gaye list ke sabhi permutations ke liye, har consecutive pair ke beech ke absolute difference ka average nikalo. \par
Args: \par
\hspace{1em}numbers (list): Sankhyaon ki ek list. \par
Returns: \par
\hspace{1em}float: Sabhi permutations ke absolute antar ka average. \par
"""
\end{minipage}
\\
\bottomrule
\end{tabular}
\end{table}

\subsection{CodeMix injection algorithm}
\label{appendix:cmd-algorithm}

The following pseudocode describes how we apply controlled code-mixing at a specified CMD level.

\begin{algorithm}
\caption{CMD-Based word replacement  
(adapted from Nayak and Joshi~\cite{nayak-joshi-2022-l3cube})}
\label{alg:cmd-replace}
\begin{algorithmic}[1]
\Require Translated sentence $S$, sorted list \texttt{sorted\_words} of tuples $(\text{eng},\text{lan},\text{score})$, code-mix degree CMD in $[0,1]$
\Ensure Code-mixed sentence $S^*$

\State $S^* \gets S$
\State $words\_replaced \gets 0$
\State $k \gets \lfloor \text{CMD} \times \lvert \texttt{sorted\_words}\rvert \rfloor$

\For{$i = 0$ \textbf{to} $\lvert \texttt{sorted\_words}\rvert - 1$}
  \State $(weng,\, wlan,\, score) \gets \texttt{sorted\_words}[i]$
  \State $remaining \gets \max\bigl(0,\,k - words\_replaced\bigr)$
  \If{$remaining = 0$}
    \State \textbf{break}
  \EndIf
  \If{$score = \infty$}
    \State $S^* \gets \textsc{ReplaceWord}(S^*,\, wlan,\, weng)$
    \State $words\_replaced \gets words\_replaced + 1$
    \State \textbf{continue}
  \EndIf
  \While{$remaining > 0$}
    \State $S^* \gets \textsc{ReplaceWord}(S^*,\, wlan,\, weng)$
    \State $words\_replaced \gets words\_replaced + 1$
    \State $remaining \gets remaining - 1$
  \EndWhile
\EndFor

\State \Return $S^*$
\end{algorithmic}
\vspace{0.5em}
{\small \textit{Note:} \texttt{lan} refers to the matrix-language word corresponding to the English token \texttt{eng}.}
\end{algorithm}

\subsection{GAME validation}
\label{alg:game-validation}

\begin{tcolorbox}[colback=gray!5, colframe=gray!40!black, title=GAME validation prompt]
\footnotesize
I will give you a text in roman-\texttt{\{lan\}} and English below:- \newline
\texttt{------------------------} \newline
\texttt{\{processed\_cand\_cm\_romanized\}} \newline
\texttt{------------------------} \newline
You have to follow these steps below to translate the text into English: \newline
1. First translate the sentence exactly into non-roman \texttt{\{lan\}} representation of characters. \newline
2. Now translate the sentence into its English translation and remember it as SE. \newline
3. Now for the sentences that are English words and are common in both the SE in step 2 and the original text, POS tag them. \newline
4. Now remembering the POS tags of words, translate the original sentence into \texttt{\{lan\}}, remembering the meaning of POS words and cross-language homonyms. \newline
5. Now translate the final \texttt{\{lan\}} sentence in step 4 to its English translation. \newline

Only give me the translated sentence in step 5 as your response in this template below so that I can extract the sentence using regular expression: \newline
\texttt{*****Final Translated Sentence*****} \newline
\texttt{[put the final sentence from step 5.]} \newline
\texttt{*****End*****}
\end{tcolorbox}

\small{\textit{Note}: The variable \texttt{lan} denotes the target matrix language, and \texttt{processed\_cand\_cm\_romanized} refers to the Roman-script code-mixed prompt being validated.}

\subsection{LLM prompt templates}
\label{appendix:prompt-templates}

\subsubsection*{A.1 Prompt for matrix language translation and dictionary construction}

\begin{tcolorbox}[colback=gray!5, colframe=gray!40!black, title=Exact prompt used]
\footnotesize
Translate this "instruct" sentence in \texttt{\{lan\}} \newline
\texttt{-------------------------------------} \newline
\texttt{\{prompt['instruct\_prompt']\}} \newline
\texttt{-------------------------------------} \newline
Do not translate the code and programming terms (args, list etc) in the prompt. Make it more like human written. \newline
And here is some of the important English PoS tags: \texttt{\{i\_imp\_eng\}} \newline
Look for the corresponding meaning of these PoS in \texttt{\{lan\}} and look for the \texttt{\{lan\}} word in the translated sentence. \newline
Now translate each \texttt{\{lan\}} word in the dictionary in Roman \texttt{\{lan\}} in three ways or spellings, all must be strictly different in spelling. Remember that the translation must be only in Anglo-Saxon script. \newline
Create a JSON dictionary which contains the eng PoS word as \texttt{eng\_word} and the corresponding \texttt{\{lan\}} word in the translated sentence as \texttt{\{lan\}\_word} if exists in \texttt{\{lan\}}, and the Roman Anglo-Saxon script translations of the \texttt{\{lan\}\_word}. \newline
Format above as RFC8259-compliant JSON dictionary, in the format: \newline
\texttt{["eng": <eng\_word>, "\{lan\}": <\{lan\}\_word>, "roman\_\{lan\}": <transliterations>]} \newline

Translate this "doc\_struct" prompt in \texttt{\{lan\}} \newline
\texttt{-------------------------------------} \newline
\texttt{\{prompt['doc\_struct']\}} \newline
\texttt{-------------------------------------} \newline
These are the requirements:- \newline
Do not translate the keys and do not translate the word 'Args'. \newline
Do not translate the code. \newline
Do not translate programming terms like args, list etc in the prompt. \newline
Do not translate any abbreviation. \newline
Do not translate 'reqs','raises','examples'. \newline
Return the whole prompt. \newline
Do not miss to include any docstring element. \newline
Only output the docstring dictionary with no other text at all. \newline
And here is some of the important English PoS tags for this sentence: \texttt{\{d\_imp\_eng\}} \newline
Look for the corresponding meaning of these PoS in \texttt{\{lan\}} and look for the \texttt{\{lan\}} word in the translated sentence. \newline
Now transliterate each \texttt{\{lan\}} word in the dictionary in Roman \texttt{\{lan\}} in three ways or spellings, all must be strictly different in spelling. Remember that the translation must be only in Anglo-Saxon script. \newline
Create a JSON dictionary which contains the eng PoS word as \texttt{eng\_word} and the corresponding \texttt{\{lan\}} word in the translated sentence as \texttt{\{lan\}\_word} if exists in \texttt{\{lan\}}, and the Roman Anglo-Saxon script translations of the \texttt{\{lan\}\_word}. \newline
Format above as RFC8259-compliant JSON dictionary, in the format: \newline
\texttt{["eng": <eng\_word>, "\{lan\}": <\{lan\}\_word>, "roman\_\{lan\}": <transliterations>]} \newline

Only return the translated prompt and the Roman dictionary in this format structure and nothing else:- \newline
\texttt{*****translated\_instruct\_prompt} \newline
\texttt{[translated\_prompt]} \newline
\texttt{*****roman\_instruct\_dictionary} \newline
\texttt{[roman\_dictionary]} \newline
\texttt{*****translated\_doc\_struct\_prompt} \newline
\texttt{[translated\_prompt]} \newline
\texttt{*****roman\_doc\_struct\_dictionary} \newline
\texttt{[roman\_dictionary]}
\end{tcolorbox}

\subsubsection*{A.2 Prompt for romanization of translated tokens}

\begin{tcolorbox}[colback=gray!5, colframe=gray!40!black, title=Romanization prompt]
\footnotesize
I want you to romanize the following from \texttt{\{lan\}} to its roman translation while keeping the words from English as it is. \newline
Remember that the translation must be only in roman Anglo-Saxon script. The translated prompt should not contain any \texttt{\{lan\}} words and all the spellings for English words must be correct. \newline
\texttt{-------------------------------------} \newline
\texttt{\{prompt\}} \newline
\texttt{-------------------------------------} \newline
Only return the roman translation in the exact format and structure as the original prompt.
\end{tcolorbox}

\subsection{Prompt examples across CMD levels}
\label{appendix:prompt_formats_cmd}

\begin{table}[h]
\caption{Prompt translations across CMD levels for a realistic logging-related task. Differences between the CMD levels—especially extra English tokens retained in CMD = 0.9—are highlighted in \textbf{bold}.}
\label{tab:cmd_prompt_examples}
\centering
\renewcommand{\arraystretch}{1.2}
\begin{tabular}{p{0.25\linewidth} p{0.7\linewidth}}
\toprule
\textbf{Prompt level} & \textbf{Prompt text} \\
\midrule
\textbf{English (original)} & 
\parbox[t]{\linewidth}{\raggedright
\footnotesize
Find the latest log file in a specified directory that matches a given regex pattern. This function searches through all files in the specified directory, filters them based on the provided regex pattern, and returns the path to the most recent log file based on modification time. If no files match the pattern or the directory is empty, the function returns None.\\

The function should output with:\\
\texttt{str or None}: The path to the most recent log file that matches the pattern, or None if no matching files are found.\\

You should write self-contained code starting with:\\
\texttt{import os}\newline
\texttt{import re}\newline
\texttt{def task\_func(pattern, log\_dir='/var/log/'):}
}
\\
\midrule
\textbf{CMD = 0.6 (Hinglish)} & 
\parbox[t]{\linewidth}{\raggedright
\footnotesize
Ek diye gaye regex pattern se mel khaane vaali, ek specified directory mein sabse \textit{latest} \textit{log} \textit{file} dhoondhe. Yeh \textit{function} specified \textit{directory} mein sabhi \textit{files} ko khojata hai, unhein pradaan kiye gaye \textit{regex} \textit{pattern} ke aadhaar par filter karta hai, \textit{and} \textit{modification} \textit{time} ke aadhaar par sabse \textit{recent} \textit{log} \textit{file} ke \textit{path} ko lautaata hai. Yadi \textit{pattern} se koi \textit{files} mel nahi khaate \textit{or} \textit{directory} \textit{empty} hai, to \textit{function} \textit{None} lautaata hai.\\

Function ko output karna chahiye:\\
\texttt{str or None}: Sabse recent log file ka path jo pattern se mel khaata hai, or None yadi koi matching files nahi paaye jaate hain.\\

Aapko yahaan se shuru hone vaala self-contained code likhna chahiye:\\
\texttt{import os}\newline
\texttt{import re}\newline
\texttt{def task\_func(pattern, log\_dir='/var/log/'):}
}
\\
\midrule
\textbf{CMD = 0.9 (Hinglish)} & 
\parbox[t]{\linewidth}{\raggedright
\footnotesize
Ek specified directory mein latest log file khojen jo diye gaye regex pattern se \textbf{matching} khati hai. Yeh function specified directory mein sabhi files ki khoj karta hai, unhein pradaan kiye gaye regex pattern ke aadhaar par filter karta hai, and modification time ke aadhaar par \textbf{most} recent \textit{log} \textit{file} ka path lautata hai. Yadi koi bhi file pattern se \textbf{matching} nahi khati hai or directory \textbf{khali} hai, to function None lautata hai.\\

Function ko output karna chahiye:\\
\texttt{str or None}: \textbf{Most} recent log file ka path jo pattern se \textbf{matching} khata hai, or yadi koi matching files nahi mili to \textbf{None}.\\

Aapko is tarah se self-contained code likhna chahiye:\\
\texttt{import os}\newline
\texttt{import re}\newline
\texttt{def task\_func(pattern, log\_dir='/var/log/'):}
}
\\
\bottomrule
\end{tabular}
\end{table}

\subsection{Prompt split examples}
\label{appendix:prompt_formats_dataset}

\begin{table}[h]
\caption{Example prompt formats from the complete and instruct splits of BigCodeBench.}
\label{tab:prompt_splits}
\centering
\renewcommand{\arraystretch}{1.1}
\begin{tabular}{p{0.47\linewidth} p{0.47\linewidth}}
\toprule
\textbf{Complete split} & \textbf{Instruct split} \\
\midrule
\begin{minipage}[t]{\linewidth}
\footnotesize
\ttfamily
\raggedright
import numpy as np\par
import math\par
\mbox{}\par
def task\_func(data, target, k):\par
\hspace{2em}"""\par
\hspace{2em}Calculate the 'k' nearest neighbors\par 
\hspace{2em}by geographic coordinates using a\par
\hspace{2em}dataset and a target data point.\par
\mbox{}\par
\hspace{2em}The function returns a list of the\par
\hspace{2em}'k' nearest neighbors, sorted in\par 
\hspace{2em}ascending order of their distances\par 
\hspace{2em}from the target.\par
\mbox{}\par
\hspace{2em}Parameters:\par
\hspace{2em}data (DataFrame): The dataset\par
\hspace{2em}containing geographical coordinates\par
\hspace{2em}with columns ['Latitude',\par
\hspace{2em}'Longitude'].\par
\hspace{2em}target (list): The target data\par
\hspace{2em}point as [Latitude, Longitude].\par
\hspace{2em}k (int): The number of nearest\par
\hspace{2em}neighbors to return. Must be a\par
\hspace{2em}non-negative integer.\par
\mbox{}\par
\hspace{2em}Returns:\par 
\hspace{2em}list: List of the 'k' nearest\par 
\hspace{2em}neighbors as [Latitude, Longitude].\par
\mbox{}\par
\hspace{2em}Raises:\par
\hspace{2em}ValueError: If 'k' is negative or\par 
\hspace{2em}not an integer\par
\mbox{}\par
\hspace{2em}Constants:\par
\hspace{2em}Radius of Earth = 6371 km\par
\mbox{}\par
\hspace{2em}Example:\par
\hspace{2em}>>> data = pd.DataFrame([\par
\hspace{2em}[14, 25], [1, 22], [7, 8]],\par
\hspace{2em}columns=['Latitude', 'Longitude'])\par
\hspace{2em}>>> target = [10, 15]\par
\hspace{2em}>>> k = 2\par
\hspace{2em}>>> task\_func(data, target, k)\par
\hspace{2em}[[7, 8], [14, 25]]\par
\hspace{2em}"""
\end{minipage}
&
\begin{minipage}[t]{\linewidth}
\footnotesize
\ttfamily
\raggedright
Calculate the 'k' nearest neighbors by geographic coordinates using a dataset and a target data point. The function returns a list of the 'k' nearest neighbors, sorted in ascending order of their distances from the target.\par
Constants: radius of earth is 6371 km.\par
The function should raise the exception for:\par
ValueError: If 'k' is a negative integer or not an integer.\par
The function should output with:\par
list: List of the 'k' nearest neighbors as [Latitude, Longitude].\par
You should write self-contained code starting with:\par
\texttt{\char96\char96\char96python}\par
import numpy as np\par
import math\par
def task\_func(data, target, k):\par
\texttt{\char96\char96\char96}\par
\end{minipage}
\\
\bottomrule
\end{tabular}
\end{table}

\end{document}